\documentclass[11pt]{article}

\PassOptionsToPackage{dvipsnames}{xcolor}

\usepackage[]{acl}

\usepackage{times}
\usepackage{latexsym}

\usepackage{lipsum}
\usepackage{arydshln}

\usepackage[utf8]{inputenc}
\usepackage[T1]{fontenc}
\usepackage{microtype}

\usepackage{inconsolata}

\usepackage{enumitem}
\usepackage[normalem]{ulem} 
\usepackage{booktabs}
\usepackage{url}
\usepackage[most]{tcolorbox}
\usepackage{fancyvrb}
\usepackage{pifont} 
\usepackage{graphicx} 
\usepackage{soul} 

\title{NADI 2025:\\ The First Multidialectal Arabic Speech Processing Shared Task}
\author{
\normalsize 
\textbf{Bashar Talafha}$^\lambda$ 
~ \textbf{Hawau Olamide Toyin}$^\xi$
~ \textbf{Peter Sullivan}$^\lambda$ 
~ \textbf{AbdelRahim Elmadany}$^\lambda$ \\ 
\normalsize 
\textbf{Abdurrahman Juma}$^\gamma$ 
~ \textbf{Amirbek Djanibekov}$^\xi$ 
~ \textbf{Chiyu Zhang}$^\lambda$
~ \textbf{Hamad Alshehhi}$^\xi$ \\
\normalsize 
~ \textbf{Hanan Aldarmaki}$^\xi$
~ \textbf{Mustafa Jarrar}$^\alpha$$^\gamma$
~ \textbf{Nizar Habash}$^\eta$$^\xi$
~ \textbf{Muhammad Abdul-Mageed}$^\lambda$ ~ \\ 
\normalsize 
$^\lambda$The University of British Columbia ~ $^\alpha$Hamad Bin Khalifa University ~
\\
\normalsize 
$^\gamma$Birzeit University ~
\normalsize  
$^\xi$MBZUAI ~ 
\normalsize  
$^\eta$NYU Abu Dhabi \\ 
\texttt{
  \small  
  \{btalafha@mail.,a.elmadany@,muhammad.mageed@\}ubc.ca} 
}

\begin{document}
\maketitle
\begin{abstract}
We present the findings of the \texttt{sixth} Nuanced Arabic Dialect Identification (\texttt{NADI $2025$}) Shared Task, which focused on Arabic speech dialect processing across three subtasks: spoken dialect identification (Subtask~1), speech recognition (Subtask~2), and diacritic restoration for spoken dialects (Subtask~3). A total of $44$ teams registered, and during the testing phase, $100$ valid submissions were received from eight unique teams. The distribution was as follows: $34$ submissions for Subtask~1 ``\textit{five teams}'', $47$ submissions for Subtask~2 ``\textit{six teams}'', and $19$ submissions for Subtask~3 ``\textit{two teams}''. The best-performing systems achieved $79.8\%$ accuracy on Subtask~1, $35.68/12.20$ WER/CER (overall average) on Subtask~2, and $55/13$ WER/CER on Subtask~3. These results highlight the ongoing challenges of Arabic dialect speech processing, particularly in dialect identification, recognition, and diacritic restoration. We also summarize the methods adopted by participating teams and briefly outline directions for future editions of NADI.
\footnote{The official leaderboards and datasets for NADI~2025 are available at 
\url{https://nadi.dlnlp.ai/2025/}}
\end{abstract}
\section{Introduction}

\begin{figure}[t]
\centering
\includegraphics[width=0.5\textwidth]{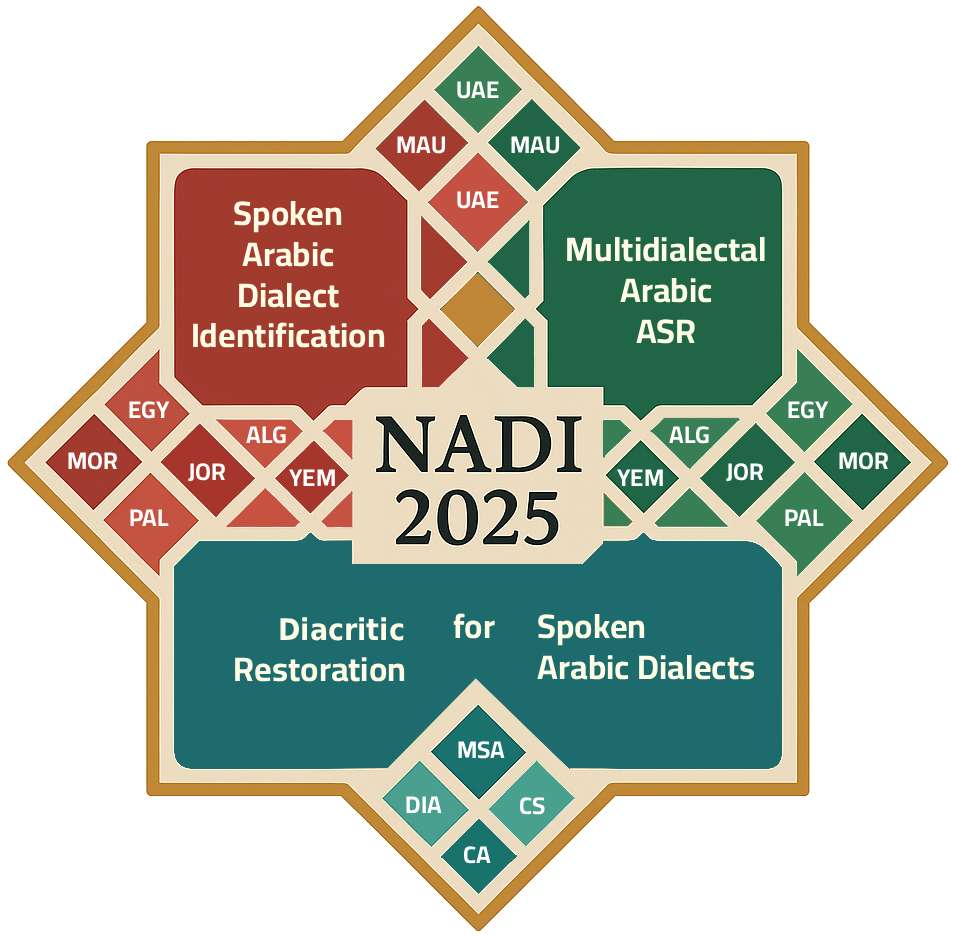}
\caption{Overview of the \textbf{NADI~2025} shared tasks.}
\label{fig:nadi_overview}
\end{figure} 

Spoken Arabic exhibits a remarkable degree of linguistic diversity. Beyond Modern Standard Arabic (MSA) and Classical Arabic (CA), which have historically dominated computational work, Arabic encompasses numerous regional and national dialects that differ across all linguistic levels (phonology, morphology, lexicon, and syntax) and in discourse/pragmatics~\citep{talafha2024casablanca,JZHNW23}. These varieties also frequently exhibit intra- and inter-sentential code-switching with other languages~\citep{abdul2024nadi}. These varieties dominate everyday communication across the Arab world yet remain under‑represented in annotated datasets and resources~\cite{madar-2018-bouamor,DH21,abdulmageed2020b, abdul2023nadi}.  At the same time, many downstream applications—from automatic transcription and virtual assistants to text‑to‑speech and educational tools—depend on accurate handling of dialectal speech and the diacritics that indicate short vowels and phonological features.  Existing systems trained on CA/MSA~\cite{elmadany2023octopus,toyin2023artst} often ignore these diacritics or assume text forms, leaving a large gap between technology and real‑world usage.


NADI shared task series, hosted at the ArabicNLP conference\footnote{Formerly the Workshop on Arabic Natural Language Processing, WANLP} since $2020$, was created to alleviate this bottleneck by providing curated datasets and standardized evaluation settings for dialect identification, translation and related tasks~\citep{abdul2020nadi,abdul2021nadi,abdul2022nadi,abdul2023nadi,abdul2024nadi}. These earlier, text‑focused editions—together with the general observation that Arabic dialects remain under‑studied due to limited resources—motivate a shift in NADI 2025 toward speech and diacritization.

NADI $2025$ marks the \textit{sixth} edition of the NADI shared task series, hosted by the Third Arabic Natural Language Processing Conference (ArabicNLP 2025\footnote{\href{https://arabicnlp2025.sigarab.org/}{https://arabicnlp2025.sigarab.org/}}). In the following, we introduce several key new features that set it apart from previous versions, focusing on the challenges of real-world, spoken Arabic dialects:

\noindent \textbf{A unified speech processing benchmark.} This edition brings together three distinct but complementary tasks, ``\textit{dialect identification}'', ``\textit{automatic speech recognition}'', and ``\textit{diacritic restoration}'', under one umbrella. This creates a comprehensive benchmark for evaluating system performance across the full spectrum of challenges in Arabic speech processing.

\noindent \textbf{New evaluation datasets and unified benchmarking framework.}
We introduce a comprehensive suite of newly-curated datasets across all three subtasks. This includes a high-quality blind test sets \textit{eight-hours} speech corpus for spoken dialect identification, a large-scale \textit{$10,807$} \textit{utterances} for ASR, and a \textit{$1,332$} utterances for diacritic restoration, all covering diverse Arabic varieties. Beyond the data itself, NADI 2025 establishes a robust and unified evaluation framework featuring large-scale blind test sets to ensure fair comparison. This framework introduces novel paradigms, such as benchmarking model \textit{adaptation} in the ADI task and offering distinct \textit{open} and \textit{closed} tracks for Diacritic Restoration. 

\noindent \textbf{A novel diacritic restoration task.} We introduce the first shared task for diacritic restoration that moves beyond formal written Arabic (CA and MSA) to target  \textit{spoken dialects} and \textit{code-switched language}. The task is uniquely designed to encourage multimodal solutions that leverage both speech and text as input.

Figure~\ref{fig:nadi_overview} provides a schematic overview of the NADI 2025 shared task, illustrating its three main subtasks including \textcolor{BrickRed}{Spoken Arabic Dialect Identification}, which covers \textit{eight} regional dialects as ``\textit{Algerian}'' (ALG), ``\textit{Egyptian}'' (EGY), ``\textit{Emirati}'' (UAE), ``\textit{Jordanian}'' (JOR), ``\textit{Mauritanian}'' (MAU), ``\textit{Moroccan}'' (MOR), ``\textit{Palestinian}'' (PAL), and ``\textit{Yemeni}'' (YEM); \textcolor{OliveGreen}{Multidialectal Arabic ASR}, which targets the exact same set of dialects; and \textcolor{teal}{Diacritic Restoration for Spoken Arabic Dialects}, which encompasses MSA, mixed dialects, code-switched varieties, and CA.

The rest of the paper is organized as follows: Section~\ref{sec:lit} provides a review of related work on spoken Arabic processing and the history of the NADI shared task. In Section~\ref{sec:nadi2025}, we describe the NADI 2025 shared task in detail, including the three subtasks, their datasets, and evaluation metrics. Section~\ref{sec:teams_results} presents the results for all participating teams and baselines, followed by an overview of the submitted systems in Section~\ref{subsec:papers-desc}. We conclude the paper in Section~\ref{sec:conclusion}.

\section{Literature Review}\label{sec:lit}


Unlike previous NADI tasks that relied on text, NADI 2025 concentrates on spoken Arabic dialects. Accordingly, this section covers related work on the subtasks of spoken language identification, ASR, and diacritic restoration.  Before delving into the related work, it is useful to explore the history of NADI and its growth since its inception.

\subsection{NADI Shared Task: Origins and Growth} 

\texttt{NADI-2020}, the first NADI shared task~\cite{abdul2020nadi} involved two subtasks, one targeting country level (21 countries) and another focusing on province level (100 provinces), both exploiting X, \textit{formerly Twitter}, data. NADI 2020 was the first shared task to exploit naturally occurring fine-grained dialectal text at the sub-country level.

\texttt{NADI-2021}, the second version~\cite{abdul2021nadi} targeted the same 21 Arab countries and 100 corresponding provinces as NADI 2020, also using X data. However, it improved upon the previous version by removing non-Arabic data and distinguishing between MSA and dialectical Arabic (DA). It involved four subtasks: MSA-country, DA-country, MSA-province, and DA-province.

\texttt{NADI-2022}~\cite{abdul2022nadi} continued the focus on studying Arabic dialects at the country level, but also included dialectal sentiment analysis with an objective to explore variation in socio-geographical regions that had not been extensively studied before.

\texttt{NADI-2023}, the fourth edition~\cite{abdul2023nadi}, proposed new machine translation subtasks from four dialectal Arabic varieties to MSA, in two themes (open-track and closed-track) as well as a dialect identification subtask at the country level. 

Finally, \texttt{NADI-2024}, the fifth edition~\cite{abdul2024nadi}, targeted both dialect ID cast as a multi-label task, identification of the Arabic level of dialectness, and dialect-to-MSA machine translation.

\subsection{Spoken Dialect Identification}
Although CA and MSA have been extensively examined \cite{harrell1962,badawi1973mustawayat,brustad2000,holes2004}, dialectal Arabic (DA) became the center of attention only relatively recently. A significant challenge in studying DA has been the scarcity of resources, prompting researchers to create new DA datasets targeting limited regions \cite{gadalla1997,diab2010,alsabbagh2012,sadat2014,harrat2014,jarrar2016,khalifa2016,altwairesh2018,alsarsour2018,kwaik2018,elhaj2020,EJHZ22,ANMFTM23,JZHNW23}. Several works introducing multi-dialectal datasets and models for region-level dialect identification \cite{zaidan2011,elfardy2014,bouamor2014,meftouh2015} and the VarDial workshop series employing transcriptions of speech broadcasts \cite{malmasi2016discriminating} also followed. Other work developed relatively small-sized commissioned data~\cite{madar-2018-bouamor,salameh2018,obeid2019}.

Subsequently, larger datasets that cover between 10 to 21 countries were introduced \cite{mubarak2014,abdulmageed2018,zaghouani2018,abdulmageed2020b,abdelali2021,issa2021,baimukan2022,althobaiti2022,elleuch25_interspeech,HKJ25}. The majority of these datasets are compiled from social media posts, especially X (formerly Twitter). More recently, benchmarks such as ORCA~\cite{elmadany2023b} and DOLPHIN~\cite{nagoudi2023} boast dialectal coverage. 

Spoken dialect ID shares with text-based dialect ID a scarcity of labeled data. Important efforts to counter this include the introduction of the multigenre and multidialectal ADI-5~\cite{ali2017speech} and ADI-17~\cite{ali2019mgb,shon2020adi17} datasets (covering coarse regional and fine-grain country-level dialects, respectively). Moving from text to speech as a modality, however, introduces additional complexities such as potential channel mismatch between train and test sets due to differences in recording conditions, as is in the case with ADI-5~\cite{ali2017speech}. Furthermore, dialect ID models may capture non-linguistic information such as gender and channel features~\cite{chowdhury2020does}, and may experience major performance degradation in cross-domain and cross-dialect settings~\cite{sullivan23_interspeech,HKJ25}.

\subsection{Automatic Speech Recognition}
Arabic ASR systems often struggle with dialectal speech, primarily due to lack of (or limited) dialectal data~\cite{waheed2023voxarabica}. Mozilla Common Voice~\cite{ardila-etal-2020-common} and MASC (MSA and Dialectal Speech)~\cite{al-fetyani2022masc} were introduced to alleviate this issue. However, both of these corpora label the data under a single label (Arabic) instead of different dialect names. Some of the audio and text samples in these datasets are also misaligned~\cite{lau-etal-2025-data}. The Casablanca project~\cite{talafha2024casablanca} compiled high quality multidialectal speech for eight countries, providing a significant boost towards research in multidialectal ASR. \citet{djanibekov2025dialectalcoveragegeneralizationarabic} have also recently presented strong results for dialectal Arabic ASR as well as training strategies that work best based on data availability for each dialect.

\subsection{Diacritic Restoration} 
Several text-based approaches~\cite{alasmary2024cattcharacterbasedarabictashkeel,elgamal-etal-2024-arabic, fadel-etal-2019-neural,harrat13_interspeech} and resources~\cite{toyin2025arvoicemultispeakerdatasetarabic,tashkeela} have been proposed for Arabic diacritic / vowel restoration. ~\citet{aldarmaki2023diacritic} found the speech based approach to outperform text only approaches. More recently, speech based or multi-modal approaches have also been proposed -albeit at a slow rate, mainly due to lack of parallel speech-text resources~\cite{shatnawi2024automaticrestorationdiacriticsspeech}.~\citet{elmadany2023octopus} report strong diacritization models as part of the Octopus toolkit, based on simple finetuning of AraT5~\cite{elmadany2022arat5}.
\citet{shatnawi2024automaticrestorationdiacriticsspeech} also propose an ASR-based diacritic restoration framework, where a pretrained ASR model generates vowelized transcripts refined by a secondary diacritization model. While their approach achieved high accuracy for CA, it fails to generalize to dialectal Arabic due to dataset limitations.

\section{NADI 2025}\label{sec:nadi2025}
NADI~2025 is the sixth edition of NADI shared task series. Since we extend the scope of the shared task to address broader challenges in multidialectal Arabic speech processing, we refer to NADI~2025 as ``\textit{the first multidialectal Arabic speech processing shared task}''. This edition comprises three complementary subtasks: spoken Arabic dialect identification, multidialectal Arabic ASR, and diacritization restoration. Collectively, these subtasks target critical components of the Arabic speech technology pipeline, each addressing long-standing challenges arising from the language's rich dialectal variation, frequent code-switching, and the absence of diacritics in most written Arabic. By curating diverse, high quality datasets and establishing standardized evaluation protocols, NADI~2025 aims to catalyze the development of robust, generalizable systems that advance state of the art in Arabic speech and language processing.

\subsection{Subtask 1 - Spoken Dialect Identification} 

\paragraph{Task Description.} This subtask is an 8-way classification task to identify which of country-level dialect is being spoken in an utterance, with our set of countries being \textit{Algeria, Egypt, Jordan, Mauritania, Morocco, Palestine, United Arab Emirates (UAE), and Yemen}.

\paragraph{Data.} In this subtask, we follow similar procedure in selecting utterances to the data collection procedure of Casablanca ~\cite{talafha2024casablanca}. For each dialect, different series were identified and the dialect spoken was verified by fluent speakers. For the Adaptation set, we utilize the same series as in Casablanca, but ensure there is no overlap with the series used for the Test set. By doing so, we aim to minimize the influence of potentially overlapping speakers, and to try to disentangle the dialect ID task from simple domain classification.

\paragraph{Evaluation Metric.} We use both accuracy as well as the Language Recognition Evaluation 2022 average Cost metric ($C_{avg}$) ~\cite{lee2022nist}. Because Cost is based on the probability of missed detections as well as false alarms for a given system it provides a complementary way to characterize model performance. At a high level, for two models that have similar accuracy but different Cost, the lower Cost model will providing a larger positive margin between the probability of the correct classes in comparison to incorrect classes, while the higher Cost model would have a smaller margin between correct and incorrect class probabilities.

\subsection{Subtask 2 - Multidialectal Arabic ASR}
\paragraph{Task Description.} The ASR subtask2 in NADI-2025 focuses on building speech recognition systems that can handle spoken Arabic across a range of regional dialects: \textit{Algerian, Egyptian, Jordanian, Mauritanian, Moroccan, Palestinian, Emirati, and Yemeni}. The task includes both monolingual and code-switched speech, which captures the variation speakers naturally use in different settings. 

\paragraph{Data.} The dataset used in this subtask is a subset of the Casablanca corpus \cite{talafha2024casablanca}. In this subtask, we select balanced samples from each dialect. The training set is intended primarily for adaptation rather than full model training, encouraging participants to leverage transfer learning, domain adaptation, and other data-efficient strategies. We provide a total of $47,027$ utterances, evenly distributed across the eight dialects for the training, validation, and test sets ($1,600$ utterances per dialect per split). The only exceptions are Algeria, Palestine, and Yemen, which have $727$, $900$, and $1,180$ utterances, respectively, in the test set. These lower counts are due to the limited availability of samples for these dialects in the original Casablanca dataset.


\paragraph{Evaluation Metric.} System performance is evaluated using the word error rate (WER) as the primary metric, reported both overall and per dialect. We also report character error rate (CER)\footnote{In the case of a tie, we use the average CER as the tiebreaker.} for additional insight into system performance, particularly for short utterances and morphologically rich forms. During evaluation, in line with~\citealp{talafha2024casablanca}, we apply a consistent text normalization pipeline to both system outputs and reference transcripts. Specifically, we:  
(a) retain only the \% symbol, removing other special characters, (b) eliminate diacritics, (c) normalize Hamzas and Maddas to bare alif,   
(d) convert Eastern Arabic numerals to Western Arabic numerals, and  
(e) preserve all Latin characters, as Casablanca contains code-switching segments in other languages.


\paragraph{Subtask 3 - Diacritic Restoration for Spoken Arabic Varieties.} This subtask aims to advance research on automatic diacritic restoration for spoken Arabic varieties. As the vast majority of existing vowelization or diacritic restoration efforts focus on CA or MSA, we aim to raise attention to more challenging spoken varieties, such as dialects and code-switching, with a focus on generalization across different varieties. The objective of this subtask  is to restore the diacritics of a given text. The text can be in a variety of forms, including MSA and Arabic dialects and may even include code-switched instances. In addition to text, all inputs have an associated speech utterance to encourage multi-modal approaches.

\paragraph{Data.} This subtask encourages the development of multi-modal (speech + text) diacritic restoration models that generalize across Arabic variants. To enable the development of such models, we identified several high-quality data sets \cite{6487288,abdallah2023leveragingdatacollectionunsupervised,al-ali-aldarmaki-2024-mixat,hamed-etal-2020-arzen,kulkarni2023clartts,toyin2025arvoicemultispeakerdatasetarabic} of Arabic variants (CA, MSA, dialectal, CS) that include parallel speech and text. \noindent Table \ref{tab:vorest-datasets} shows a summary of the data sets provided to the participants for this subtask. The MDASPC dataset contains multi-dialectal speech with diacritized transcriptions and we include it for training. For the \textit{TunSwitch}~\cite{abdallah2023leveragingdatacollectionunsupervised} training data, we used GPT-4o with a chain-of-thought prompt to initially diacritize the transcriptions. The diacritized output of GPT-4o was subsequently manually corrected with the corresponding audio as a reference by a native Arabic speaker. For code-switching, we provide undiacritized resources for training; \textit{ArzEn}~\cite{hamed-etal-2020-arzen}, \textit{Mixat}~\cite{al-ali-aldarmaki-2024-mixat} and \textit{MGB2}~\cite{7846277}; for each dataset, we provide diacritized test sets by manually annotating random subsets of their test set. 
\begin{table}
\resizebox{\columnwidth}{!}{%
\begin{tabular}{@{}lllrrr@{}}
\toprule
\textbf{Dataset}   & \textbf{Type}            & \textbf{Diacritized} & \textbf{Train} & \textbf{Dev}  & \textbf{Test \textbf{(Ours)}} \\ \midrule
MDASPC      & Multi-dialectal & True        & $60,677$ & ---    & $5,164$            \\
TunSwitch & Dialectal, CS   & True       & $5,212$  & $165$  & $110$ (\textbf{110})              \\
ArzEn     & Dialectal, CS   & False       & $3,344$  & $1,402$ & $1,470$ (\textbf{104})          \\
Mixat     & Dialectal, CS   & False       & $3,721$  & ---    & $1,583$ (\textbf{100})          \\
ClArTTS   & CA              & True        & $9,500$  & ---    & 204                \\ 
ArVoice & MSA& True & $2,507$ & $258$ & (\textbf{11})  \\ 
MGB2 & MSA & False & --- & --- & $5,365$ (\textbf{40}) \\
\bottomrule
\end{tabular}%
}
\caption{Number of sentences in datasets provided for the diacritic restoration sub-task. \textbf{{Ours}.} refers to the held-out test set for this shared task which we manually diaritize. \textbf{CA.} refers to Classical Arabic. \textbf{CS.} refers to code-switching. 
}
\label{tab:vorest-datasets}
\end{table}

\paragraph{Evaluation Metric.} Similar to subtask 2, we use WER and CER as performance metrics for this subtask, which are chosen to enable the evaluation of diacritic restoration performance even for models that may change the underlying text, such as ASR-based or sequence-to-sequence models. 

\section{Shared Task Teams \& Results}\label{sec:teams_results}

\begin{table*}[!ht]
\centering
\resizebox{0.8\textwidth}{!}{%
\begin{tabular}{ll c}
\toprule
\textbf{Team Name} & \textbf{Affiliation} & \textbf{Subtask}  \\
\midrule
\textbf{Abjad AI}~\cite{abjad-dr} & Abjad AI, Jordan \& Saudi Arabia & 1,3 \\
\textbf{BYZÖ}~\cite{byzo} & Saarland University, Germany & 2  \\
\textbf{Elyadata}~\cite{elyadata} & Elyadata, Tunisia &1,2 \\
\textbf{Hamsa} & Hamsa & 2 \\
\textbf{Lahjati}~\cite{lahjati} & Princess Sumaya University, Jordan &1  \\
\textbf{MarsadLab}~\cite{marsad-lab} & Hamad Bin Khalifa University, Qatar & 1,2  \\
\textbf{Munsit}~\cite{munist} & Lebanese American University, Lebanon &2  \\
\textbf{Unicorn}~\cite{unicorn} & Ain shams University, Egypt&3 \\
\bottomrule
\end{tabular}
}
\caption{List of teams that participated in NADI 2025 shared task. Teams with accepted papers are cited.}
\label{tab:teams}
\end{table*}
\subsection{Participating Teams}
A total of $44$ teams registered for the NADI 2025. At the testing phase, a total of $100$ valid entries were submitted by \textit{eight} unique teams. The breakdown across the subtasks as follow: $34$ submissions for subtask 1 by \textit{five} teams, $47$ submissions for subtask 2 by \textit{six} teams and $19$ submissions by \textit{two} teams for subtask 3. Table \ref{tab:teams} list NADI 2025 participated teams which completed the testing phase.


\subsection{Baselines}
We developed baseline (BL) models for each subtask to serve as reference points for evaluating the teams’ systems. These models were not shared with participants during the competition.

\paragraph{Subtask 1.} We finetune SpeechBrain's VoxLingua107~\cite{valk2021voxlingua107} ECAPA-TDNN ~\cite{desplanques20_interspeech} system\footnote{\url{https://huggingface.co/speechbrain/spkrec-ecapa-voxceleb}} on the adaptation split of the dataset. We replace the classification layers of the pretrained system with new randomized layers corresponding to the smaller number of output classes ($8$); and train these new layers with the rest of the model frozen for $5$K steps, and then unfreeze the model and train for an additional $25$K steps. We use AdamW with base learning rate of $1e-4$, and apply a linear ramp up from $1/3$ base LR over $3$K steps followed by constant LR until unfreezing, and then repeat the linear ramp up and plateau. Finally, Starting at $20$K steps we applying an exponential decay.

\paragraph{Subtask 2.} A zero-shot baseline is built on the pre-trained Whisper-Large-v3 model~\cite{radford2022whisper}. Dialect-wise inference is performed on the official NADI 2025 subtask~2 ASR release available on Hugging Face\footnote{\url{https://huggingface.co/datasets/UBC-NLP/NADI2025_subtask2_ASR}}, which provides validation splits for eight country-level dialects; official evaluation is conducted on a private Codabench test set. During inference, audio inputs are transcribed using Whisper's default decoding parameters with language explicitly set to Arabic.

\begin{table}
\centering
\resizebox{0.9\columnwidth}{!}{%
\begin{tabular}{lcc}
\toprule
 & \textbf{Accuracy} $\uparrow$ & \textbf{C}$_{avg}\downarrow$ \\
\midrule
\textbf{ELYDATA-LIA}~\cite{elyadata} & \textbf{79.8} & \textbf{17.88} \\
\textbf{BYZÖ-ADI}~\cite{byzo} & 76.4 &  22.65  \\
\textbf{MarsadLab}~\cite{marsad-lab} & 61.6 &   30.68 \\
\textbf{Abjad AI} & 61.2 &  34.77  \\
\cdashline{1-3}
\textbf{Baseline} & 61.1 &   34.22 \\
\cdashline{1-3}
\textbf{Lahjati}~\cite{lahjati} & 50.8 &   48.99 \\
\bottomrule

\end{tabular}
}
\caption{Performance of the systems on the test set for Subtask 1. Results are sorted by Accuracy, while the average cost (C$_{avg}$) score is also reported, with lower values indicating better performance. The best performance is highlighted in bold.}
\label{tab:ADI-res}
\end{table}

\paragraph{Subtask 3.} In this subtask, we provide three baselines: (I) A \textit{text only} baseline based on the publicly available CATT model~\cite{alasmary2024cattcharacterbasedarabictashkeel}, which we use without further fine-tuning (II) an \textit{ASR based} baseline where we use the ArTST v3 checkpoint \cite{djanibekov2025dialectalcoveragegeneralizationarabic}, which is pre-trained on dialectal and code-switched Arabic, and finetune it for Arabic ASR with diacritics using the provided training data, and (III) a \textit{multi-modal} diacritc restoration model designed as follows: 

The raw waveform and corresponding transcriptions are passed in parallel to a speech and text encoder, respectively. The speech encoder is derived from ArTST ASR ~\cite{djanibekov2025dialectalcoveragegeneralizationarabic}, and the text encoder is derived from ArTST TTS model~\cite{toyin2023artst}. We then align the resulting text and speech embeddings using multi-head attention with $8$ heads, 
followed by a trainable prediction component comprising $2$ bi-directional LSTM layers, a $3$0\% dropout layer, and a final linear prediction head to predict the corresponding diacritics. Simple ad-hoc post-processing is applied to add the predicted diacritics to the input text to produce the fully diacritized text output. This approach is inspired by the multi-modal diacritization model described in \citet{shatnawi2024automaticrestorationdiacriticsspeech}.


\begin{table*}

\resizebox{\textwidth}{!}{%
\begin{tabular}{lccccccccc}
\toprule
\multicolumn{1}{l}{ }& \multicolumn{1}{c}{\textbf{Average}}                        & \multicolumn{1}{c}{\textbf{JOR}}                       & \multicolumn{1}{c}{\textbf{EGY}}                        & \multicolumn{1}{c}{\textbf{MOR}}                        & \multicolumn{1}{c}{\textbf{ALG}}                        & \multicolumn{1}{l}{\textbf{YEM}}                        & \multicolumn{1}{l}{\textbf{MAU}}                        & \multicolumn{1}{c}{\textbf{UAE}}                        & \multicolumn{1}{c}{\textbf{PAL}}                        \\ \midrule
\textbf{Munsit}~\cite{munist}                   & \begin{tabular}[c]{@{}c@{}}\textbf{35.68}/\textbf{12.20}\end{tabular}  & \begin{tabular}[c]{@{}c@{}}\textbf{20.68}/\textbf{5.64}\end{tabular}  & \begin{tabular}[c]{@{}c@{}}\textbf{20.88}/\textbf{7.33}\end{tabular}   & \begin{tabular}[c]{@{}c@{}}41.71/14.04\end{tabular}  & \begin{tabular}[c]{@{}c@{}}\textbf{53.62}/\textbf{18.44}\end{tabular}  & \begin{tabular}[c]{@{}c@{}}\textbf{44.62}/\textbf{14.30}\end{tabular}  & \begin{tabular}[c]{@{}c@{}}59.03/\textbf{23.28}\end{tabular}  & \begin{tabular}[c]{@{}c@{}}\textbf{22.66}/\textbf{6.55}\end{tabular}   & \begin{tabular}[c]{@{}c@{}}\textbf{22.27}/\textbf{8.05}\end{tabular}   \\
\textbf{ELYADATA-LIA}~\cite{elyadata}              & \begin{tabular}[c]{@{}c@{}}38.53/14.52\end{tabular}  & \begin{tabular}[c]{@{}c@{}}28.03/9.36\end{tabular}  & \begin{tabular}[c]{@{}c@{}}26.83/11.43\end{tabular}  & \begin{tabular}[c]{@{}c@{}}\textbf{38.26}/\textbf{13.66}\end{tabular}  & \begin{tabular}[c]{@{}c@{}}53.73/20.43\end{tabular}  & \begin{tabular}[c]{@{}c@{}}46.63/16.66\end{tabular}  & \begin{tabular}[c]{@{}c@{}}\textbf{58.10}/24.53\end{tabular}  & \begin{tabular}[c]{@{}c@{}}29.35/9.91\end{tabular}   & \begin{tabular}[c]{@{}c@{}}27.36/10.20\end{tabular}  \\
\textbf{BYZÖ-Whisper}~\cite{byzo}                     & \begin{tabular}[c]{@{}c@{}}39.78/14.75\end{tabular}  & \begin{tabular}[c]{@{}c@{}}28.84/9.47\end{tabular}  & \begin{tabular}[c]{@{}c@{}}29.50/11.91\end{tabular}  & \begin{tabular}[c]{@{}c@{}}43.06/15.52\end{tabular}  & \begin{tabular}[c]{@{}c@{}}55.04/20.59\end{tabular}  & \begin{tabular}[c]{@{}c@{}}46.41/16.05\end{tabular}  & \begin{tabular}[c]{@{}c@{}}59.36/24.84\end{tabular}  & \begin{tabular}[c]{@{}c@{}}28.38/9.04\end{tabular}   & \begin{tabular}[c]{@{}c@{}}27.65/10.59\end{tabular}  \\
\textbf{Hamsa}              & \begin{tabular}[c]{@{}c@{}}42.04/16.18\end{tabular}  & \begin{tabular}[c]{@{}c@{}}32.24/9.90\end{tabular}  & \begin{tabular}[c]{@{}c@{}}24.72/10.21\end{tabular}  & \begin{tabular}[c]{@{}c@{}}48.21/18.11\end{tabular}  & \begin{tabular}[c]{@{}c@{}}60.32/23.33\end{tabular}  & \begin{tabular}[c]{@{}c@{}}51.76/20.41\end{tabular}  & \begin{tabular}[c]{@{}c@{}}66.23/29.11\end{tabular}  & \begin{tabular}[c]{@{}c@{}}28.00/8.98\end{tabular}   & \begin{tabular}[c]{@{}c@{}}24.87/9.41\end{tabular}   \\
\textbf{BYZÖ-CTC}~\cite{byzo}                & \begin{tabular}[c]{@{}c@{}}44.14/15.58\end{tabular}  & \begin{tabular}[c]{@{}c@{}}31.74/9.94\end{tabular}  & \begin{tabular}[c]{@{}c@{}}37.23/12.57\end{tabular}  & \begin{tabular}[c]{@{}c@{}}43.31/15.07\end{tabular}  & \begin{tabular}[c]{@{}c@{}}56.12/21.38\end{tabular}  & \begin{tabular}[c]{@{}c@{}}46.14/15.68\end{tabular}  & \begin{tabular}[c]{@{}c@{}}63.32/26.70\end{tabular}  & \begin{tabular}[c]{@{}c@{}}38.65/11.14\end{tabular}  & \begin{tabular}[c]{@{}c@{}}36.62/12.18\end{tabular}  \\
\cdashline{1-10}
\textbf{Baseline}                    & \begin{tabular}[c]{@{}c@{}}93.89/72.79\end{tabular}  & \begin{tabular}[c]{@{}c@{}}46.09/19.28\end{tabular} & \begin{tabular}[c]{@{}c@{}}100.06/81.37\end{tabular} & \begin{tabular}[c]{@{}c@{}}100.38/80.42\end{tabular} & \begin{tabular}[c]{@{}c@{}}101.03/79.58\end{tabular} & \begin{tabular}[c]{@{}c@{}}101.09/80.58\end{tabular} & \begin{tabular}[c]{@{}c@{}}100.59/82.89\end{tabular} & \begin{tabular}[c]{@{}c@{}}101.15/80.27\end{tabular} & \begin{tabular}[c]{@{}c@{}}100.76/77.92\end{tabular} \\
\cdashline{1-10}
\textbf{MarsadLab}~\cite{marsad-lab}                & \begin{tabular}[c]{@{}c@{}}104.89/84.69\end{tabular} & \begin{tabular}[c]{@{}c@{}}44.97/19.19\end{tabular} & \begin{tabular}[c]{@{}c@{}}113.97/97.65\end{tabular} & \begin{tabular}[c]{@{}c@{}}104.07/87.58\end{tabular} & \begin{tabular}[c]{@{}c@{}}116.59/94.26\end{tabular} & \begin{tabular}[c]{@{}c@{}}113.54/94.56\end{tabular} & \begin{tabular}[c]{@{}c@{}}111.59/92.84\end{tabular} & \begin{tabular}[c]{@{}c@{}}116.79/97.00\end{tabular} & \begin{tabular}[c]{@{}c@{}}117.60/94.42\end{tabular} \\ \bottomrule
\end{tabular}
}

\caption{Performance of the systems on the test set for Subtask 2. Results are sorted by the overall average WER/CER score across all dialects, with lower values indicating better performance. The best performance is highlighted in bold.}
\label{tab:ASR-res}
\end{table*}
\subsection{Results}

Tables~~\ref{tab:ADI-res}, \ref{tab:ASR-res}, and \ref{tab:DR-res}, present the preformernce of the submitted systems on the test set for subtask 1, subtask 2, and subtask 3 respectively. 



\paragraph{Subtask1.} The \texttt{ELYDATA-LIA} team~\cite{elyadata} achieved the best performance in terms of both accuracy and average cost $C_{avg}$ (79.8 / 17.88), followed closely by \texttt{BYZÖ-ADI}~\cite{byzo} (76.4 / 22.65). Both top teams addressed the limited size of the Adaptation set in novel ways: ELYDATA-LIA leveraged the much larger ADI-20 dataset~\cite{elleuch25_interspeech}, while BYZÖ-ADI employed kNN voice conversion~\cite{baas23_interspeech} to augment the training data with synthetic samples. In third place, \texttt{MarsadLab}~\cite{marsad-lab} improved upon the baseline system through additional data augmentation and the introduction of an attention mechanism prior to the classification layer. In fourth place, \texttt{Abjad AI} fine-tuned a Whisper Small encoder with further data augmentation. While the third- and fourth-place systems were close in terms of accuracy (61.6 vs. 61.2), the approach by MarsadLa achieved a notably better $C_{avg}$, reducing it by approximately 4 points. Finally, we note that one team (Lahjati~\cite{lahjati}) perform below the baseline. Overall, these results highlight the effectiveness and diversity of data augmentation strategies.

\paragraph{Subtask 2.} The \texttt{Munist} team~\cite{munist} obtain the lowest overall average WER/CER scores (35.68/12.10) among all participating systems, achieving the best performance across all dialects except Moroccan, where it ranked second in both WER and CER, and Mauritanian, where it ranked first in CER and second in WER. The \texttt{ELYADATA-LIA} team~\cite{elyadata} ranked second with scores of 38.52/14.52. They achieved the best performance on the Moroccan dialect (WER/CER of 38.26/13.66) and obtain the lowest CER for the Mauritanian dialect. Their performance on the Algerian dialect was only marginally lower than that of the first-ranked team, suggesting that their system demonstrates strong capabilities for North African dialects in general. The \texttt{BYZÖ-Whisper} team~\cite{byzo} ranked third, with average WER/CER scores of 39.78/14.75. The \texttt{Hamsa} team follow in fourth place, scoring 42.04/16.18, while the \texttt{BYZÖ-CTC} team~\cite{byzo} ranked fifth with 44.14/15.58. Only one team, \texttt{MarsadLab}~\cite{marsad-lab}, perform below the baseline, with notably higher average WER/CER scores of 104.89/84.69. The winning team \texttt{Munist}~\cite{munist} surpassed the baseline by \textbf{58.21} WER points (93.89 $\rightarrow$ $35.68$; $\approx$ $62$\% reduction). Furthermore, the variation in $WER$ scores among the teams that surpassed the baseline is relatively low ($\sigma \approx 3.25$), corresponding to about $8.1\%$ of the mean WER for these systems.

\paragraph{Subtask 3.} The \texttt{Abjad AI}~\cite{abjad-dr} perform the best with the lowest WER of 55\% and CER of 13\%. The \texttt{Unicorn} team~\cite{unicorn} follow closely behind with WER of 64\% and CER of 15\%. Both teams improve over the provided baselines, the best of which achieve WER and CER of 65\% and 16\%, respectively.

\begin{table}
\centering
\resizebox{0.9\columnwidth}{!}{%
\begin{tabular}{llcc}
\toprule
 & & \textbf{WER} $\downarrow$ & \textbf{CER} $\downarrow$\\
\midrule
\textbf{Abjad AI}~\cite{abjad-dr} & & \textbf{55} &  \textbf{13}  \\
\textbf{Unicorn}~\cite{unicorn} & & 64 &  15  \\
\cdashline{1-3}
\textbf{Baseline-I (\textit{ASR based})} & & 88 & 45 \\
\textbf{Baseline-II (\textit{text-only)}} & & 65 &   16 \\
\textbf{Baseline-II (\textit{multi-modal})} & & 66 & 16  \\
\bottomrule

\end{tabular}
}

\caption{Performance of the systems on the test set for Subtask 3. Results are sorted by the overall average WER/CER score across all dialects, with lower values indicating better performance. The best performance is highlighted in bold.}
\label{tab:DR-res}
\end{table}


\section{Overview of Submitted Systems}\label{subsec:papers-desc}
In this section, we present an overview of the submitted systems for each subtask and summarize the methodological approaches adopted by the participating teams.

\subsection{Subtask 1}

\paragraph{ELYDATA-LIA~\cite{elyadata}.} Using Whisper Large-v3 encoder as their base model, they adopt a two stage finetuning procedure to first finetune on the forthcoming ADI-20 dataset~\cite{elleuch25_interspeech}, and then use the NADI ADI Adaptation set for a second finetuning. Features of this approach include freezing the first 16 layers of the encoder and using plenty of data augmentation methods including speed perturbation, added noise, and frequency and chunk dropping.

\paragraph{BYZÖ-ADI~\cite{byzo}} The authors choose a straightforward finetuning approach using w2v-BERT-2.0~\cite{barrault2023seamless} model finetuned on the NADI ADI split (69\% accuracy). However, in order to improve the robustness of the model, they add a data augmentation approach by using a voice conversion model~\cite{baas23_interspeech} to re-synthesizing the training utterances using voice samples from the 4 Arabic speakers from the LibriVox project, and training on the mixed natural and synthetic audio, leading to their final model.

\paragraph{MarsadLab~\cite{marsad-lab}} Adopts a starting point similar to the baseline with a VoxLingua107 ECAPA-TDNN system that was finetuned on the ADI task. They introduce a number of features in the process including feature reweighting of the hidden representation just prior to the classification layer through the use of a lightweight attention mechanism, discriminative learning rate of the classification head, progressive unfreezing, as well as data augmentation using SpecAugment and injected noise.

\paragraph{Abjad AI} Like the ELYDATA-LIA approach, this team used Whisper model, Whisper Small, and finetuned the encoder for dialect ID. They use only the NADI Adaptation set for finetuning, using SpecAugment (time and frequency masking) for data augmentation, and unfreezing the model partway through training.

\paragraph{Lahjati~\cite{lahjati}} Using both the VoxLingua107 ECAPA-TDNN system as well as WavLM encoder, this fusion approach concattenates the outputs from the two models (WavLM pooled to match the ECAPA 256 dimension), and passes this combined representation through a layer normalization layer and then a two layers feedforward network to perform classification. Similar to other approaches the underlying models start frozen, with unfreezing at 8000 steps, followed by a ramp up, plateau, and then cosine annealing learning rate schedule.


\subsection{Subtask 2} 


\paragraph{Munsit~\cite{munist}} This system follows a two-stage training pipeline combining large-scale weakly supervised pretraining and continual supervised fine-tuning, inspired by~\citealp{salhab2025advancing}. In the first stage, a Conformer-large model~\cite{gulati2020conformer} (121M parameters) was pretrained on 15K hours of weakly labeled Arabic speech, covering MSA and various dialects, with automatic labeling and no manual verification. In the second stage, the model was fine-tuned using a high-quality dataset composed of 3,000 hours of rigorously filtered weakly labeled data, excluding news content, and the official Casablanca Challenge training set, expanded via data augmentation. Training used the CTC~\cite{graves2006connectionist} objective with a SentencePiece~\cite{kudo2018sentencepiece} vocabulary of 128 tokens, AdamW optimizer, Noam learning rate schedule, and dropout of 0.1, in a distributed setup across 8 NVIDIA A100 GPUs with bfloat16 precision. This approach enabled robust performance across all dialects, achieving the lowest average WER and CER in the shared task.


\paragraph{ELYADATA \& LIA~\cite{elyadata}} For the ASR subtask, this team fine-tuned the SeamlessM4T-v2~\cite{barrault2023seamless} Large Egyptian model separately for each of the eight dialects in the Casablanca dataset, producing eight distinct models. Training was performed for 6 epochs on NVIDIA A100 GPUs with a label-smoothed NLL loss (smoothing 0.2), AdamW optimizer, and a learning rate schedule with 100 warm-up steps ramping from 1e-9 to 5e-5. A batch size of 2 was used for all runs. This per-dialect fine-tuning approach yielded second overall in the shared task.

\paragraph{BYZÖ~\cite{byzo}} The team submitted two independent systems. The first, \texttt{BYZÖ-Whisper}, fine-tuned the Whisper-Large-v3 model~\cite{radford2023robust} (1.54B parameters) for Arabic dialect ASR using only the NADI shared task data, without external datasets or data augmentation. Text labels were preprocessed by removing bracketed content and normalizing spacing. Training followed a two-stage process: (1) domain adaptation on combined data from all dialects for 9000 steps (learning rate 1e-5, batch size 32), and (2) dialect-specific adaptation for 2000 steps per dialect using CER as the metric. The second, \texttt{BYZÖ-CTC}, fine-tuned the w2v-BERT-2.0 model~\cite{barrault2023seamless} (580M parameters) using a mix of public Arabic ASR datasets, then further fine-tuned per dialect on the shared task data (learning rate 1e-5, batch size 16). A multi-dialectal 3-gram Kneser-Ney smoothed language model, trained on collected dialect-specific text data, was integrated to reduce WER. This encoder-only CTC-based system was noted for efficiency and competitive zero-shot performance compared to Whisper large.


\paragraph{MarsadLab~\cite{marsad-lab}} For the ASR subtask, this team adopted Whisper-Large model~\cite{radford2023robust} in a zero-shot setting, without any fine-tuning, preprocessing, or post-processing. Leveraging Whisper’s multilingual capabilities, the system directly transcribed Arabic speech from multiple dialects in the test set. While the ECAPA-TDNN architecture was central to their ADI submission, it was not applied to ASR.

\paragraph{Hamsa} Submissions were received from the \texttt{Hamsa} team; however, a system description was not made available. 

\subsection{Subtask 3} 

\paragraph{Unicorn ~\cite{unicorn}} This team addressed the diacritic restoration task by fine-tuning the GEMM3N\footnote{\url{https://unsloth.ai/}} multimodal model on both audio and text inputs. They formed diacritic restoration as a structured generation task where the model receives an undiacritized sentence and its corresponding audio and generates a fully diacritized version. They fine-tuned with LoRA adaptation to efficiently adapt the model with the provided data for the sub-task only. They applied \textit{nlpaug} for speech augmentation to simulate more diverse audio inputs. They perform inference inference by prompting GEMM3N with the raw audio and the corresponding undiacritized text.

\paragraph{Abjad AI ~\cite{abjad-dr}} This team presented CATT-Whisper which is a multimodal approach that combines both textual and speech information. Their model represents the text modality using an encoder extracted from their pre-trained model named CATT~\cite{alasmary2024cattcharacterbasedarabictashkeel}. The speech component is handled by the encoder module of the OpenAI Whisper base model~\cite{radford2022whisper}. Their approach uses two integration strategies. The former consists of fusing the speech tokens with the input at an early stage where the 1500 frames of the audio segment are averaged on the basis of 10 consecutive frames resulting in 150 speech tokens only. To ensure embedding compatibility, these averaged tokens are processed through a linear projection layer prior to merging them with the text tokens. Contextual encoding is guaranteed by the CATT encoder module. The latter strategy relies on cross-attention where text and speech embeddings are fused. Then, finally the cross-attention output is fed to the CATT classification head for token-level diacritic prediction. They randomly deactivate the speech input during training for robustness, this allows the model to perform well with or without speech.

\section{Conclusion}\label{sec:conclusion}

The \textit{sixth} NADI shared task extends the scope of the series beyond text-based processing to encompass speech and diacritization, introducing three new subtasks: spoken dialect identification, Arabic ASR, and diacritic restoration. By releasing high-quality resources and providing clear evaluation protocols, our goal is to foster progress in inclusive Arabic speech processing. This edition, we received $44$ registrations, with \textit{eight} teams submitting system outputs and \textit{seven} system description papers accepted. Results across the three subtasks highlight substantial headroom for improvement: even strong pretrained models continue to face challenges with multidialectal variability, code-switching, and diacritic restoration. We hope that this edition not only advances the state of the art on each individual subtask but also inspires future research toward unified, dialect-aware speech technologies for Arabic. 
\section*{Limitations \& Ethical Considerations}\label{sec:limit}


Despite the contributions of this year’s shared task, several limitations remain across the three subtasks:

\noindent \textbf{Coverage of dialects:} Not all Arabic dialects are represented in the test sets, which limits the generalizability of results across the full dialect continuum.

\noindent \textbf{Country-level labeling:} We acknowledge that the use of country-level labels may be problematic. The continuum of Arabic dialects is complex, and using country affiliation as a stand-in for well-defined linguistic boundaries is not without limitations. This choice was made to ensure a reasonable degree of diversity in dialect coverage, while avoiding assumptions about the generalizability of models trained on a subset of dialects to unseen but related varieties.

\noindent \textbf{Code-switching:} The datasets capture only a limited subset of code-switching phenomena, whereas real-world Arabic speech often involves more diverse language mixing.

\noindent \textbf{Real-world conditions:} Background noise, spontaneous disfluencies, and accented speech are underrepresented in the datasets, limiting ecological validity.

\noindent \textbf{Evaluation metrics}: Metrics such as WER and CER may be misleading in the ASR task, since a dialectal utterance can often have multiple valid references. As the data provides only one reference per utterance, evaluation scores may underestimate system performance by penalizing alternative but correct transcriptions.
\section*{Acknowledgments}

Muhammad Abdul-Mageed acknowledges support from Canada Research Chairs (CRC), the Natural Sciences and Engineering Research Council of Canada (NSERC; RGPIN-2018-04267), the Social Sciences and Humanities Research Council of Canada (SSHRC; 435-2018-0576; 895-2020-1004; 895-2021-1008), Canadian Foundation for Innovation (CFI; 37771), Digital Research Alliance of Canada,\footnote{\href{https://alliancecan.ca}{https://alliancecan.ca}} and UBC ARC-Sockeye.

\normalem
\bibliography{custom, referece_from_nadi2024}

\begin{thebibliography}{90}
\providecommand{\natexlab}[1]{#1}

\bibitem[{Abdallah et~al.(2023)Abdallah, Kabboudi, Kanoun, and Zaiem}]{abdallah2023leveragingdatacollectionunsupervised}
Ahmed Amine~Ben Abdallah, Ata Kabboudi, Amir Kanoun, and Salah Zaiem. 2023.
\newblock \href {https://arxiv.org/abs/2309.11327} {Leveraging data collection and unsupervised learning for code-switched tunisian arabic automatic speech recognition}.
\newblock \emph{Preprint}, arXiv:2309.11327.

\bibitem[{Abdelali et~al.(2021)Abdelali, Mubarak, Samih, Hassan, and Darwish}]{abdelali2021}
Ahmed Abdelali, Hamdy Mubarak, Younes Samih, Sabit Hassan, and Kareem Darwish. 2021.
\newblock Qadi: Arabic dialect identification in the wild.
\newblock In \emph{Proceedings of the Sixth Arabic Natural Language Processing Workshop}, pages 1--10, Kyiv, Ukraine (Virtual). Association for Computational Linguistics.

\bibitem[{Abdul-Mageed et~al.(2018)Abdul-Mageed, Alhuzali, and Elaraby}]{abdulmageed2018}
Muhammad Abdul-Mageed, Hassan Alhuzali, and Mohamed Elaraby. 2018.
\newblock You tweet what you speak: A city-level dataset of arabic dialects.
\newblock In \emph{Proceedings of the Eleventh International Conference on Language Resources and Evaluation (LREC 2018)}, Miyazaki, Japan. European Language Resources Association (ELRA).

\bibitem[{Abdul-Mageed et~al.(2023)Abdul-Mageed, Elmadany, Zhang, Nagoudi, Bouamor, and Habash}]{abdul2023nadi}
Muhammad Abdul-Mageed, AbdelRahim Elmadany, Chiyu Zhang, El~Moatez~Billah Nagoudi, Houda Bouamor, and Nizar Habash. 2023.
\newblock Nadi 2023: The fourth nuanced arabic dialect identification shared task.
\newblock \emph{arXiv preprint arXiv:2310.16117}.

\bibitem[{Abdul-Mageed et~al.(2024)Abdul-Mageed, Keleg, Elmadany, Zhang, Hamed, Magdy, Bouamor, and Habash}]{abdul2024nadi}
Muhammad Abdul-Mageed, Amr Keleg, AbdelRahim Elmadany, Chiyu Zhang, Injy Hamed, Walid Magdy, Houda Bouamor, and Nizar Habash. 2024.
\newblock Nadi 2024: The fifth nuanced arabic dialect identification shared task.
\newblock \emph{arXiv preprint arXiv:2407.04910}.

\bibitem[{Abdul{-}Mageed et~al.(2020)Abdul{-}Mageed, Zhang, Bouamor, and Habash}]{abdul2020nadi}
Muhammad Abdul{-}Mageed, Chiyu Zhang, Houda Bouamor, and Nizar Habash. 2020.
\newblock \href {https://www.aclweb.org/anthology/2020.wanlp-1.9/} {{NADI} 2020: The first nuanced arabic dialect identification shared task}.
\newblock In \emph{Proceedings of the Fifth Arabic Natural Language Processing Workshop, WANLP@COLING 2020, Barcelona, Spain (Online), December 12, 2020}, pages 97--110. Association for Computational Linguistics.

\bibitem[{Abdul-Mageed et~al.(2022)Abdul-Mageed, Zhang, Elmadany, Bouamor, and Habash}]{abdul2022nadi}
Muhammad Abdul-Mageed, Chiyu Zhang, AbdelRahim Elmadany, Houda Bouamor, and Nizar Habash. 2022.
\newblock Nadi 2022: The third nuanced arabic dialect identification shared task.
\newblock \emph{arXiv preprint arXiv:2210.09582}.

\bibitem[{Abdul-Mageed et~al.(2020)Abdul-Mageed, Zhang, Elmadany, and Ungar}]{abdulmageed2020b}
Muhammad Abdul-Mageed, Chiyu Zhang, AbdelRahim Elmadany, and Lyle Ungar. 2020.
\newblock Toward micro-dialect identification in diaglossic and code-switched environments.
\newblock In \emph{Proceedings of the 2020 Conference on Empirical Methods in Natural Language Processing (EMNLP)}, pages 5855--5876, Online. Association for Computational Linguistics.

\bibitem[{Abdul{-}Mageed et~al.(2021)Abdul{-}Mageed, Zhang, Elmadany, Bouamor, and Habash}]{abdul2021nadi}
Muhammad Abdul{-}Mageed, Chiyu Zhang, AbdelRahim~A. Elmadany, Houda Bouamor, and Nizar Habash. 2021.
\newblock \href {https://www.aclweb.org/anthology/2021.wanlp-1.28/} {{NADI} 2021: The second nuanced arabic dialect identification shared task}.
\newblock In \emph{Proceedings of the Sixth Arabic Natural Language Processing Workshop, {WANLP} 2021, Kyiv, Ukraine (Virtual), April 9, 2021}, pages 244--259. Association for Computational Linguistics.

\bibitem[{Abdullah et~al.(2025)Abdullah, Al-Ghussein, Al-Khalili, Özyilmaz, Valdenegro-Toro, Ostermann, and Klakow}]{byzo}
Badr Abdullah, Yusser Al-Ghussein, Zena Al-Khalili, Ömer Özyilmaz, Matias Valdenegro-Toro, Simon Ostermann, and Dietrich Klakow. 2025.
\newblock Saarland-groningen at nadi 2025 shared task: Effective dialectal arabic speech processing under data constraints.
\newblock In \emph{The Third Arabic Natural Language Processing Conference (ArabicNLP 2025)}, Suzhou. Association for Computational Linguistics.

\bibitem[{Abu~Kwaik et~al.(2018)Abu~Kwaik, Saad, Chatzikyriakidis, and Dobnik}]{kwaik2018}
Kathrein Abu~Kwaik, Motaz Saad, Stergios Chatzikyriakidis, and Simon Dobnik. 2018.
\newblock Shami: A corpus of levantine arabic dialects.
\newblock In \emph{Proceedings of the Eleventh International Conference on Language Resources and Evaluation (LREC 2018)}.

\bibitem[{Al~Ali and Aldarmaki(2024)}]{al-ali-aldarmaki-2024-mixat}
Maryam~Khalifa Al~Ali and Hanan Aldarmaki. 2024.
\newblock \href {https://aclanthology.org/2024.sigul-1.26} {Mixat: A data set of bilingual emirati-{E}nglish speech}.
\newblock In \emph{Proceedings of the 3rd Annual Meeting of the Special Interest Group on Under-resourced Languages @ LREC-COLING 2024}, pages 222--226, Torino, Italia. ELRA and ICCL.

\bibitem[{Al-Fetyani et~al.(2022)Al-Fetyani, Al-Barham, Abandah, Alsharkawi, and Dawas}]{al-fetyani2022masc}
Mohammad Al-Fetyani, Muhammad Al-Barham, Gheith~A. Abandah, Adham Alsharkawi, and Maha Dawas. 2022.
\newblock \href {https://doi.org/10.1109/SLT54892.2023.10022652} {{MASC}: Massive arabic speech corpus}.
\newblock In \emph{Proceedings of the 2022 IEEE Spoken Language Technology Workshop (SLT)}, page 1002206.

\bibitem[{Al-Sabbagh and Girju(2012)}]{alsabbagh2012}
Rania Al-Sabbagh and Roxana Girju. 2012.
\newblock Yadac: Yet another dialectal arabic corpus.
\newblock In \emph{Proceedings of the Eighth International Conference on Language Resources and Evaluation (LREC'12)}, pages 2882--2889, Istanbul, Turkey. European Language Resources Association (ELRA).

\bibitem[{Al-Twairesh et~al.(2018)Al-Twairesh, Al-Matham, Madi, Almugren, Al-Aljmi, Alshalan, Alshalan, Alrumayyan, Al-Manea, Bawazeer, Al-Mutlaq, Almanea, Bin~Huwaymil, Alqusair, Alotaibi, Al-Senaydi, and Alfutamani}]{altwairesh2018}
Nora Al-Twairesh, Rawan~N. Al-Matham, Nora Madi, Nada Almugren, Al-Hanouf Al-Aljmi, Shahad Alshalan, Raghad Alshalan, Nafla Alrumayyan, Shams Al-Manea, Sumayah Bawazeer, Nourah Al-Mutlaq, Nada Almanea, Waad Bin~Huwaymil, Dalal Alqusair, Reem Alotaibi, Suha Al-Senaydi, and Abeer Alfutamani. 2018.
\newblock Suar: Towards building a corpus for the saudi dialect.
\newblock In \emph{Fourth International Conference on Arabic Computational Linguistics, ACLING 2018}, volume 142 of \emph{Procedia Computer Science}, pages 72--82, Dubai, United Arab Emirates. Elsevier.

\bibitem[{Alasmary et~al.(2024)Alasmary, Zaafarani, and Ghannam}]{alasmary2024cattcharacterbasedarabictashkeel}
Faris Alasmary, Orjuwan Zaafarani, and Ahmad Ghannam. 2024.
\newblock \href {https://arxiv.org/abs/2407.03236} {Catt: Character-based arabic tashkeel transformer}.
\newblock \emph{Preprint}, arXiv:2407.03236.

\bibitem[{ALBawwab and Qawasmeh(2025)}]{lahjati}
Sanad ALBawwab and Omar Qawasmeh. 2025.
\newblock Lahjati at nadi 2025 a ecapa-wavlm fusion with multi-stage optimization.
\newblock In \emph{The Third Arabic Natural Language Processing Conference (ArabicNLP 2025)}, Suzhou. Association for Computational Linguistics.

\bibitem[{Aldarmaki and Ghannam(2023)}]{aldarmaki2023diacritic}
Hanan Aldarmaki and Ahmad Ghannam. 2023.
\newblock Diacritic recognition performance in arabic asr.
\newblock In \emph{Proc. Interspeech 2023}, pages 361--365.

\bibitem[{Ali et~al.(2016)Ali, Bell, Glass, Messaoui, Mubarak, Renals, and Zhang}]{7846277}
Ahmed Ali, Peter Bell, James Glass, Yacine Messaoui, Hamdy Mubarak, Steve Renals, and Yifan Zhang. 2016.
\newblock \href {https://doi.org/10.1109/SLT.2016.7846277} {The mgb-2 challenge: Arabic multi-dialect broadcast media recognition}.
\newblock In \emph{2016 IEEE Spoken Language Technology Workshop (SLT)}, pages 279--284.

\bibitem[{Ali et~al.(2019)Ali, Shon, Samih, Mubarak, Abdelali, Glass, Renals, and Choukri}]{ali2019mgb}
Ahmed Ali, Suwon Shon, Younes Samih, Hamdy Mubarak, Ahmed Abdelali, James Glass, Steve Renals, and Khalid Choukri. 2019.
\newblock The mgb-5 challenge: Recognition and dialect identification of dialectal arabic speech.
\newblock In \emph{2019 IEEE Automatic Speech Recognition and Understanding Workshop (ASRU)}, pages 1026--1033. IEEE.

\bibitem[{Ali et~al.(2017)Ali, Vogel, and Renals}]{ali2017speech}
Ahmed Ali, Stephan Vogel, and Steve Renals. 2017.
\newblock Speech recognition challenge in the wild: Arabic mgb-3.
\newblock In \emph{2017 IEEE Automatic Speech Recognition and Understanding Workshop (ASRU)}, pages 316--322. IEEE.

\bibitem[{Almeman et~al.(2013)Almeman, Lee, and Almiman}]{6487288}
Khalid Almeman, Mark Lee, and Ali~Abdulrahman Almiman. 2013.
\newblock \href {https://doi.org/10.1109/ICCSPA.2013.6487288} {Multi dialect arabic speech parallel corpora}.
\newblock In \emph{2013 1st International Conference on Communications, Signal Processing, and their Applications (ICCSPA)}, pages 1--6.

\bibitem[{Alsarsour et~al.(2018)Alsarsour, Mohamed, Suwaileh, and Elsayed}]{alsarsour2018}
Israa Alsarsour, Esraa Mohamed, Reem Suwaileh, and Tamer Elsayed. 2018.
\newblock Dart: A large dataset of dialectal arabic tweets.
\newblock In \emph{Proceedings of the Eleventh International Conference on Language Resources and Evaluation (LREC 2018)}.

\bibitem[{Althobaiti(2022)}]{althobaiti2022}
Maha~J. Althobaiti. 2022.
\newblock Creation of annotated country-level dialectal arabic resources: An unsupervised approach.
\newblock \emph{Natural Language Engineering}, 28(5):607--648.

\bibitem[{Ardila et~al.(2020)Ardila, Branson, Davis, Kohler, Meyer, Henretty, Morais, Saunders, Tyers, and Weber}]{ardila-etal-2020-common}
Rosana Ardila, Megan Branson, Kelly Davis, Michael Kohler, Josh Meyer, Michael Henretty, Reuben Morais, Lindsay Saunders, Francis~M. Tyers, and Gregor Weber. 2020.
\newblock Common voice: A massively-multilingual speech corpus.
\newblock In \emph{Proceedings of the 12th Conference on Language Resources and Evaluation (LREC)}, pages 4218--4222.

\bibitem[{Attia et~al.(2025)Attia, Biswas, Ibrahim, Bessghaier, Alam, and Zaghouani}]{marsad-lab}
Kais Attia, Md.~Rafiul Biswas, Shimaa Ibrahim, Mabrouka Bessghaier, Firoj Alam, and Wajdi Zaghouani. 2025.
\newblock Marsadlab at nadi: Arabic dialect identification and speech recognition using ecapa-tdnn and whisper.
\newblock In \emph{The Third Arabic Natural Language Processing Conference (ArabicNLP 2025)}, Suzhou. Association for Computational Linguistics.

\bibitem[{Baas et~al.(2023)Baas, {van Niekerk}, and Kamper}]{baas23_interspeech}
Matthew Baas, Benjamin {van Niekerk}, and Herman Kamper. 2023.
\newblock \href {https://doi.org/10.21437/Interspeech.2023-419} {Voice conversion with just nearest neighbors}.
\newblock In \emph{Interspeech 2023}, pages 2053--2057.

\bibitem[{Badawi(1973)}]{badawi1973mustawayat}
As-Said~Muh{\'a}mmad Badawi. 1973.
\newblock \emph{Mustawayat al-arabiyya al-muasira fi Misr}.
\newblock Dar al-maarif.

\bibitem[{Baimukan et~al.(2022)Baimukan, Bouamor, and Habash}]{baimukan2022}
Nurpeiis Baimukan, Houda Bouamor, and Nizar Habash. 2022.
\newblock Hierarchical aggregation of dialectal data for arabic dialect identification.
\newblock In \emph{Proceedings of the Thirteenth Language Resources and Evaluation Conference (LREC 2022)}, pages 4586--4596, Marseille, France. European Language Resources Association (ELRA).

\bibitem[{Barrault et~al.(2023)Barrault, Chung, Meglioli, Dale, Dong, Duppenthaler, Duquenne, Ellis, Elsahar, Haaheim et~al.}]{barrault2023seamless}
Lo{\"\i}c Barrault, Yu-An Chung, Mariano~Coria Meglioli, David Dale, Ning Dong, Mark Duppenthaler, Paul-Ambroise Duquenne, Brian Ellis, Hady Elsahar, Justin Haaheim, and 1 others. 2023.
\newblock Seamless: Multilingual expressive and streaming speech translation.
\newblock \emph{arXiv preprint arXiv:2312.05187}.

\bibitem[{Bouamor et~al.(2014)Bouamor, Habash, and Oflazer}]{bouamor2014}
Houda Bouamor, Nizar Habash, and Kemal Oflazer. 2014.
\newblock A multidialectal parallel corpus of arabic.
\newblock In \emph{Proceedings of the Ninth International Conference on Language Resources and Evaluation (LREC'14)}, pages 1240--1245, Reykjavik, Iceland. European Language Resources Association (ELRA).

\bibitem[{Bouamor et~al.(2018)Bouamor, Habash, Salameh, Zaghouani, Rambow, Abdulrahim, Obeid, Khalifa, Eryani, Erdmann, and Oflazer}]{madar-2018-bouamor}
Houda Bouamor, Nizar Habash, Mohammad Salameh, Wajdi Zaghouani, Owen Rambow, Dana Abdulrahim, Ossama Obeid, Salam Khalifa, Fadhl Eryani, Alexander Erdmann, and Kemal Oflazer. 2018.
\newblock \href {http://www.lrec-conf.org/proceedings/lrec2018/summaries/351.html} {The {MADAR} arabic dialect corpus and lexicon}.
\newblock In \emph{Proceedings of the Eleventh International Conference on Language Resources and Evaluation, {LREC} 2018, Miyazaki, Japan, May 7-12, 2018}. European Language Resources Association {(ELRA)}.

\bibitem[{Brustad(2000)}]{brustad2000}
Kristen Brustad. 2000.
\newblock \emph{The Syntax of Spoken Arabic: A Comparative Study of Moroccan, Egyptian, Syrian, and Kuwaiti Dialects}.
\newblock Georgetown University Press.

\bibitem[{Chowdhury et~al.(2020)Chowdhury, Ali, Shon, and Glass}]{chowdhury2020does}
Shammur~A Chowdhury, Ahmed Ali, Suwon Shon, and James~R Glass. 2020.
\newblock What does an end-to-end dialect identification model learn about non-dialectal information?
\newblock In \emph{INTERSPEECH}, pages 462--466.

\bibitem[{Darwish et~al.(2021)Darwish, Habash, Abbas, Al-Khalifa, Al-Natsheh, Bouamor, Bouzoubaa, Cavalli-Sforza, El-Beltagy, El-Hajj, Jarrar, and Mubarak}]{DH21}
Kareem Darwish, Nizar Habash, Mourad Abbas, Hend Al-Khalifa, Huseein~T. Al-Natsheh, Houda Bouamor, Karim Bouzoubaa, Violetta Cavalli-Sforza, Samhaa~R. El-Beltagy, Wassim El-Hajj, Mustafa Jarrar, and Hamdy Mubarak. 2021.
\newblock \href {https://doi.org/10.1145/3447735} {{A} {P}anoramic survey of {N}atural {L}anguage {P}rocessing in the {A}rab {W}orlds}.
\newblock \emph{Commun. ACM}, 64(4):72–81.

\bibitem[{Desplanques et~al.(2020)Desplanques, Thienpondt, and Demuynck}]{desplanques20_interspeech}
Brecht Desplanques, Jenthe Thienpondt, and Kris Demuynck. 2020.
\newblock \href {https://doi.org/10.21437/Interspeech.2020-2650} {Ecapa-tdnn: Emphasized channel attention, propagation and aggregation in tdnn based speaker verification}.
\newblock In \emph{Interspeech 2020}, pages 3830--3834.

\bibitem[{Diab et~al.(2010)Diab, Habash, Rambow, Altantawy, and Benajiba}]{diab2010}
Mona Diab, Nizar Habash, Owen Rambow, Mohamed Altantawy, and Yassine Benajiba. 2010.
\newblock Colaba: Arabic dialect annotation and processing.
\newblock In \emph{Proceedings of the LREC Workshop on Semitic Language Processing}, pages 66--74.

\bibitem[{Djanibekov et~al.(2025)Djanibekov, Toyin, Alshalan, Alitr, and Aldarmaki}]{djanibekov2025dialectalcoveragegeneralizationarabic}
Amirbek Djanibekov, Hawau~Olamide Toyin, Raghad Alshalan, Abdullah Alitr, and Hanan Aldarmaki. 2025.
\newblock \href {https://arxiv.org/abs/2411.05872} {Dialectal coverage and generalization in arabic speech recognition}.
\newblock \emph{Preprint}, arXiv:2411.05872.

\bibitem[{El-Haj(2020)}]{elhaj2020}
Mahmoud El-Haj. 2020.
\newblock Habibi -- a multi dialect multi national arabic song lyrics corpus.
\newblock In \emph{Proceedings of the 12th Language Resources and Evaluation Conference}, pages 1318--1326, Marseille, France. European Language Resources Association (ELRA).

\bibitem[{Elfardy et~al.(2014)Elfardy, Al-Badrashiny, and Diab}]{elfardy2014}
Heba Elfardy, Mohamed Al-Badrashiny, and Mona Diab. 2014.
\newblock Aida: Identifying code switching in informal arabic text.
\newblock In \emph{Proceedings of the First Workshop on Computational Approaches to Code Switching}, pages 94--101, Doha, Qatar. Association for Computational Linguistics.

\bibitem[{Elgamal et~al.(2024)Elgamal, Obeid, Kabbani, Inoue, and Habash}]{elgamal-etal-2024-arabic}
Salman Elgamal, Ossama Obeid, Mhd Kabbani, Go~Inoue, and Nizar Habash. 2024.
\newblock \href {https://doi.org/10.18653/v1/2024.acl-long.792} {{A}rabic diacritics in the wild: Exploiting opportunities for improved diacritization}.
\newblock In \emph{Proceedings of the 62nd Annual Meeting of the Association for Computational Linguistics (Volume 1: Long Papers)}, pages 14815--14829, Bangkok, Thailand. Association for Computational Linguistics.

\bibitem[{Elleuch et~al.({2025})Elleuch, Mdhaffar, Estève, and Bougares}]{elleuch25_interspeech}
Haroun Elleuch, Salima Mdhaffar, Yannick Estève, and Fethi Bougares. {2025}.
\newblock \href {https://doi.org/{10.21437/Interspeech.2025-884}} {{ADI-20: Arabic Dialect Identification dataset and models}}.
\newblock In \emph{{Interspeech 2025}}, pages {2775--2779}.

\bibitem[{Elleuch et~al.(2025)Elleuch, Saidi, Mdhaffar, Estève, and Bougares}]{elyadata}
Haroun Elleuch, Youssef Saidi, Salima Mdhaffar, Yannick Estève, and Fethi Bougares. 2025.
\newblock Elyadata \& lia at nadi 2025: Asr and adi subtasks.
\newblock In \emph{The Third Arabic Natural Language Processing Conference (ArabicNLP 2025)}, Suzhou. Association for Computational Linguistics.

\bibitem[{Elmadany et~al.(2022)Elmadany, Abdul-Mageed et~al.}]{elmadany2022arat5}
AbdelRahim Elmadany, Muhammad Abdul-Mageed, and 1 others. 2022.
\newblock Arat5: Text-to-text transformers for arabic language generation.
\newblock In \emph{Proceedings of the 60th annual meeting of the association for computational linguistics (Volume 1: Long papers)}, pages 628--647.

\bibitem[{Elmadany et~al.(2023{\natexlab{a}})Elmadany, Abdul-Mageed et~al.}]{elmadany2023octopus}
Abdelrahim Elmadany, Muhammad Abdul-Mageed, and 1 others. 2023{\natexlab{a}}.
\newblock Octopus: A multitask model and toolkit for arabic natural language generation.
\newblock In \emph{Proceedings of ArabicNLP 2023}, pages 232--243.

\bibitem[{Elmadany et~al.(2023{\natexlab{b}})Elmadany, Nagoudi, and Abdul-Mageed}]{elmadany2023b}
AbdelRahim Elmadany, ElMoatez~Billah Nagoudi, and Muhammad Abdul-Mageed. 2023{\natexlab{b}}.
\newblock Orca: A challenging benchmark for arabic language understanding.
\newblock In \emph{Findings of the Association for Computational Linguistics: ACL 2023}, pages 9559--9586, Toronto, Canada. Association for Computational Linguistics.

\bibitem[{Elrefai(2025)}]{unicorn}
Mohamed~Lotfy Elrefai. 2025.
\newblock Unicorn at nadi 2025 subtask 3: Gemm3n-dr: Audio-text diacritic restoration via fine-tuned multimodal arabic llm.
\newblock In \emph{The Third Arabic Natural Language Processing Conference (ArabicNLP 2025)}, Suzhou. Association for Computational Linguistics.

\bibitem[{Fadel et~al.(2019)Fadel, Tuffaha, Al-Jawarneh, and Al-Ayyoub}]{fadel-etal-2019-neural}
Ali Fadel, Ibraheem Tuffaha, Bara{'} Al-Jawarneh, and Mahmoud Al-Ayyoub. 2019.
\newblock \href {https://doi.org/10.18653/v1/D19-5229} {Neural {A}rabic text diacritization: State of the art results and a novel approach for machine translation}.
\newblock In \emph{Proceedings of the 6th Workshop on Asian Translation}, pages 215--225, Hong Kong, China. Association for Computational Linguistics.

\bibitem[{Gadalla et~al.(1997)Gadalla, Kilany, Arram, Yacoub, El-Habashi, Shalaby, Karins, Rowson, MacIntyre, Kingsbury, Graff, and McLemore}]{gadalla1997}
Hassan Gadalla, Hanaa Kilany, Howaida Arram, Ashraf Yacoub, Alaa El-Habashi, Amr Shalaby, Krisjanis Karins, Everett Rowson, Robert MacIntyre, Paul Kingsbury, David Graff, and Cynthia McLemore. 1997.
\newblock Callhome egyptian arabic transcripts ldc97t19.
\newblock Web Download.

\bibitem[{Ghannam et~al.(2025)Ghannam, Alharthi, Alasmary, Al~Tabash, Sadah, and Ghouti}]{abjad-dr}
Ahmad Ghannam, Naif Alharthi, Faris Alasmary, Kholood Al~Tabash, Shouq Sadah, and Lahouari Ghouti. 2025.
\newblock Abjad ai at nadi 2025: Catt-whisper: Multimodal diacritic restoration using text and speech representations.
\newblock In \emph{The Third Arabic Natural Language Processing Conference (ArabicNLP 2025)}, Suzhou. Association for Computational Linguistics.

\bibitem[{Graves et~al.(2006)Graves, Fern{\'a}ndez, Gomez, and Schmidhuber}]{graves2006connectionist}
Alex Graves, Santiago Fern{\'a}ndez, Faustino Gomez, and J{\"u}rgen Schmidhuber. 2006.
\newblock Connectionist temporal classification: labelling unsegmented sequence data with recurrent neural networks.
\newblock In \emph{Proceedings of the 23rd international conference on Machine learning}, pages 369--376.

\bibitem[{Gulati et~al.(2020)Gulati, Qin, Chiu, Parmar, Zhang, Yu, Han, Wang, Zhang, Wu et~al.}]{gulati2020conformer}
Anmol Gulati, James Qin, Chung-Cheng Chiu, Niki Parmar, Yu~Zhang, Jiahui Yu, Wei Han, Shibo Wang, Zhengdong Zhang, Yonghui Wu, and 1 others. 2020.
\newblock Conformer: Convolution-augmented transformer for speech recognition.
\newblock \emph{arXiv preprint arXiv:2005.08100}.

\bibitem[{Haff et~al.(2022)Haff, Jarrar, Hammouda, and Zaraket}]{EJHZ22}
Karim~El Haff, Mustafa Jarrar, Tymaa Hammouda, and Fadi Zaraket. 2022.
\newblock \href {https://aclanthology.org/2022.lrec-1.82.pdf} {{C}urras + {B}aladi: {T}owards a {L}evantine {C}orpus}.
\newblock In \emph{Proceedings of the International Conference on Language Resources and Evaluation(LREC 2022)}, Marseille, France.

\bibitem[{Hamad et~al.(2025)Hamad, Khalilia, and Jarrar}]{HKJ25}
Nagham Hamad, Mohammed Khalilia, and Mustafa Jarrar. 2025.
\newblock \href {http://www.jarrar.info/publications/HKJ25.pdf} {{K}onooz: {M}ulti-domain {M}ulti-dialect {C}orpus for {N}amed {E}ntity {R}ecognition}.
\newblock In \emph{Proceedings of the 63rd Annual Meeting of the Association for Computational Linguistics}, pages 0--0, Vienna, Austria. Association for Computational Linguistics.

\bibitem[{Hamed et~al.(2020)Hamed, Vu, and Abdennadher}]{hamed-etal-2020-arzen}
Injy Hamed, Ngoc~Thang Vu, and Slim Abdennadher. 2020.
\newblock \href {https://aclanthology.org/2020.lrec-1.523/} {{A}rz{E}n: A speech corpus for code-switched {E}gyptian {A}rabic-{E}nglish}.
\newblock In \emph{Proceedings of the Twelfth Language Resources and Evaluation Conference}, pages 4237--4246, Marseille, France. European Language Resources Association.

\bibitem[{Harrat et~al.(2013)Harrat, Abbas, Meftouh, and Smaili}]{harrat13_interspeech}
S.~Harrat, M.~Abbas, K.~Meftouh, and K.~Smaili. 2013.
\newblock \href {https://doi.org/10.21437/Interspeech.2013-373} {Diacritics restoration for arabic dialect texts}.
\newblock In \emph{Interspeech 2013}, pages 1429--1433.

\bibitem[{Harrat et~al.(2014)Harrat, Meftouh, Abbas, and Sma{\"i}li}]{harrat2014}
Salima Harrat, Karima Meftouh, Mourad Abbas, and Kamel Sma{\"i}li. 2014.
\newblock Building resources for algerian arabic dialects.
\newblock In \emph{INTERSPEECH 2014, 15th Annual Conference of the International Speech Communication Association}, pages 2123--2127, Singapore. ISCA.

\bibitem[{Harrell(1962)}]{harrell1962}
Richard~S. Harrell. 1962.
\newblock \emph{A Short Reference Grammar of Moroccan Arabic: With Audio CD}.
\newblock Georgetown Classics in Arabic Language and Linguistics. Georgetown University Press.

\bibitem[{Holes(2004)}]{holes2004}
Clive Holes. 2004.
\newblock \emph{Modern Arabic: Structures, Functions, and Varieties}.
\newblock Georgetown Classics in Arabic Language and Linguistics. Georgetown University Press.

\bibitem[{Issa et~al.(2021)Issa, AlShakhori, AlBahrani, and Hahn-Powell}]{issa2021}
Elsayed Issa, Mohammed AlShakhori, Reda AlBahrani, and Gus Hahn-Powell. 2021.
\newblock Country-level arabic dialect identification using rnns with and without linguistic features.
\newblock In \emph{Proceedings of the Sixth Arabic Natural Language Processing Workshop}, pages 276--281, Kyiv, Ukraine (Virtual). Association for Computational Linguistics.

\bibitem[{Jarrar et~al.(2016)Jarrar, Habash, Alrimawi, Akra, and Zalmout}]{jarrar2016}
Mustafa Jarrar, Nizar Habash, Faeq Alrimawi, Diyam Akra, and Nasser Zalmout. 2016.
\newblock Curras: An annotated corpus for the palestinian arabic dialect.
\newblock \emph{Language Resources and Evaluation}, pages 1--31.

\bibitem[{Jarrar et~al.(2023)Jarrar, Zaraket, Hammouda, Alavi, and Waahlisch}]{JZHNW23}
Mustafa Jarrar, Fadi Zaraket, Tymaa Hammouda, Daanish~Masood Alavi, and Martin Waahlisch. 2023.
\newblock \href {https://doi.org/1 10.1109/AICCSA59173.2023.10479250} {{L}isan: {Y}emeni, {I}rqi, {L}ibyan, and {S}udanese {A}rabic {D}ialect {C}opora with {M}orphological {A}nnotations}.
\newblock In \emph{The 20th IEEE/ACS International Conference on Computer Systems and Applications (AICCSA)}, pages 1--7. IEEE.

\bibitem[{Khalifa et~al.(2016)Khalifa, Habash, Abdulrahim, and Hassan}]{khalifa2016}
Salam Khalifa, Nizar Habash, Dana Abdulrahim, and Sara Hassan. 2016.
\newblock A large scale corpus of gulf arabic.
\newblock In \emph{Proceedings of the Tenth International Conference on Language Resources and Evaluation (LREC'16)}, pages 4282--4289, Portorož, Slovenia. European Language Resources Association (ELRA).

\bibitem[{Kudo and Richardson(2018)}]{kudo2018sentencepiece}
Taku Kudo and John Richardson. 2018.
\newblock Sentencepiece: A simple and language independent subword tokenizer and detokenizer for neural text processing.
\newblock \emph{arXiv preprint arXiv:1808.06226}.

\bibitem[{Kulkarni et~al.(2023)Kulkarni, Kulkarni, Shatnawi, and Aldarmaki}]{kulkarni2023clartts}
Ajinkya Kulkarni, Atharva Kulkarni, Sara Abedalmon'em~Mohammad Shatnawi, and Hanan Aldarmaki. 2023.
\newblock \href {https://doi.org/10.21437/Interspeech.2023-2224} {Clartts: An open-source classical arabic text-to-speech corpus}.
\newblock In \emph{2023 INTERSPEECH}, pages 5511--5515.

\bibitem[{Lau et~al.(2025)Lau, Chen, Fang, Xu, Chen, and Golik}]{lau-etal-2025-data}
Mingfei Lau, Qian Chen, Yeming Fang, Tingting Xu, Tongzhou Chen, and Pavel Golik. 2025.
\newblock \href {https://doi.org/10.18653/v1/2025.acl-long.370} {Data quality issues in multilingual speech datasets: The need for sociolinguistic awareness and proactive language planning}.
\newblock In \emph{Proceedings of the 63rd Annual Meeting of the Association for Computational Linguistics (Volume 1: Long Papers)}, pages 7466--7492, Vienna, Austria. Association for Computational Linguistics.

\bibitem[{Lee et~al.(2022)Lee, Greenberg, Mason, and Singer}]{lee2022nist}
Yooyoung Lee, Craig Greenberg, Lisa Mason, and Elliot Singer. 2022.
\newblock Nist 2022 language recognition evaluation plan.

\bibitem[{Malmasi et~al.(2016)Malmasi, Zampieri, Ljube{\v{s}}i{\'c}, Nakov, Ali, and Tiedemann}]{malmasi2016discriminating}
Shervin Malmasi, Marcos Zampieri, Nikola Ljube{\v{s}}i{\'c}, Preslav Nakov, Ahmed Ali, and J{\"o}rg Tiedemann. 2016.
\newblock Discriminating between similar languages and {A}rabic dialect identification: A report on the third {DSL} shared task.
\newblock In \emph{Proceedings of the third workshop on NLP for similar languages, varieties and dialects (VarDial3)}, pages 1--14.

\bibitem[{Meftouh et~al.(2015)Meftouh, Harrat, Jamoussi, Abbas, and Smaili}]{meftouh2015}
Karima Meftouh, Salima Harrat, Salma Jamoussi, Mourad Abbas, and Kamel Smaili. 2015.
\newblock Machine translation experiments on padic: A parallel arabic dialect corpus.
\newblock In \emph{Proceedings of the 29th Pacific Asia Conference on Language, Information and Computation}, pages 26--34, Shanghai, China.

\bibitem[{Mubarak and Darwish(2014)}]{mubarak2014}
Hamdy Mubarak and Kareem Darwish. 2014.
\newblock Using twitter to collect a multi-dialectal corpus of arabic.
\newblock In \emph{Proceedings of the EMNLP 2014 Workshop on Arabic Natural Language Processing (ANLP)}, pages 1--7, Doha, Qatar. Association for Computational Linguistics.

\bibitem[{Nagoudi et~al.(2023)Nagoudi, El-Shangiti, Elmadany, and Abdul-Mageed}]{nagoudi2023}
El~Moatez~Billah Nagoudi, Ahmed El-Shangiti, AbdelRahim Elmadany, and Muhammad Abdul-Mageed. 2023.
\newblock Dolphin: A challenging and diverse benchmark for arabic nlg.
\newblock \emph{arXiv preprint arXiv:2305.14989}.

\bibitem[{Nayouf et~al.(2023)Nayouf, Jarrar, zaraket, Hammouda, and Kurdy}]{ANMFTM23}
Amal Nayouf, Mustafa Jarrar, Fadi zaraket, Tymaa Hammouda, and Mohamad-Bassam Kurdy. 2023.
\newblock \href {https://doi.org/10.18653/v1/2023.arabicnlp-1.2} {{N}âbra: {S}yrian {A}rabic {D}ialects with {M}orphological {A}nnotations}.
\newblock In \emph{Proceedings of the 1st Arabic Natural Language Processing Conference (ArabicNLP), Part of the EMNLP 2023}, pages 12--23. ACL.

\bibitem[{Obeid et~al.(2019)Obeid, Salameh, Bouamor, and Habash}]{obeid2019}
Ossama Obeid, Mohammad Salameh, Houda Bouamor, and Nizar Habash. 2019.
\newblock Adida: Automatic dialect identification for arabic.
\newblock In \emph{Proceedings of the 2019 Conference of the North American Chapter of the Association for Computational Linguistics (Demonstrations)}, pages 6--11, Minneapolis, Minnesota. Association for Computational Linguistics.

\bibitem[{Radford et~al.(2022)Radford, Kim, Xu, Brockman, McLeavey, and Sutskever}]{radford2022whisper}
Alec Radford, Jong~Wook Kim, Tao Xu, Greg Brockman, Christine McLeavey, and Ilya Sutskever. 2022.
\newblock \href {https://doi.org/10.48550/ARXIV.2212.04356} {Robust speech recognition via large-scale weak supervision}.
\newblock \emph{arXiv preprint}.

\bibitem[{Radford et~al.(2023)Radford, Kim, Xu, Brockman, McLeavey, and Sutskever}]{radford2023robust}
Alec Radford, Jong~Wook Kim, Tao Xu, Greg Brockman, Christine McLeavey, and Ilya Sutskever. 2023.
\newblock Robust speech recognition via large-scale weak supervision.
\newblock In \emph{International conference on machine learning}, pages 28492--28518. PMLR.

\bibitem[{Sadat et~al.(2014)Sadat, Kazemi, and Farzindar}]{sadat2014}
Fatiha Sadat, Farzindar Kazemi, and Atefeh Farzindar. 2014.
\newblock Automatic identification of arabic language varieties and dialects in social media.
\newblock In \emph{Proceedings of the Second Workshop on Natural Language Processing for Social Media (SocialNLP)}, pages 22--27, Dublin, Ireland. Association for Computational Linguistics and Dublin City University.

\bibitem[{Salameh et~al.(2018)Salameh, Bouamor, and Habash}]{salameh2018}
Mohammad Salameh, Houda Bouamor, and Nizar Habash. 2018.
\newblock Fine-grained arabic dialect identification.
\newblock In \emph{Proceedings of the 27th International Conference on Computational Linguistics}, pages 1332--1344, Santa Fe, New Mexico, USA. Association for Computational Linguistics.

\bibitem[{Salhab et~al.(2025{\natexlab{a}})Salhab, Elghitany, Sait, Ullah, Abusheikh, and Abusheikh}]{salhab2025advancing}
Mahmoud Salhab, Marwan Elghitany, Shameed Sait, Syed~Sibghat Ullah, Mohammad Abusheikh, and Hasan Abusheikh. 2025{\natexlab{a}}.
\newblock Advancing arabic speech recognition through large-scale weakly supervised learning.
\newblock \emph{arXiv preprint arXiv:2504.12254}.

\bibitem[{Salhab et~al.(2025{\natexlab{b}})Salhab, Sait, Abusheikh, and Abusheikh}]{munist}
Mahmoud Salhab, Shameed Sait, Mohammad Abusheikh, and Hasan Abusheikh. 2025{\natexlab{b}}.
\newblock Munsit at nadi 2025 shared task 2: Pushing the boundaries of multidialectal arabic asr with weakly supervised pretraining and continual supervised fine-tuning.
\newblock In \emph{The Third Arabic Natural Language Processing Conference (ArabicNLP 2025)}, Suzhou. Association for Computational Linguistics.

\bibitem[{Shatnawi et~al.(2024)Shatnawi, Alqahtani, and Aldarmaki}]{shatnawi2024automaticrestorationdiacriticsspeech}
Sara Shatnawi, Sawsan Alqahtani, and Hanan Aldarmaki. 2024.
\newblock Automatic restoration of diacritics for speech data sets.
\newblock In \emph{Proceedings of the 2024 Conference of the North American Chapter of the Association for Computational Linguistics: Human Language Technologies (Volume 1: Long Papers)}, pages 4166--4176.

\bibitem[{Shon et~al.(2020)Shon, Ali, Samih, Mubarak, and Glass}]{shon2020adi17}
Suwon Shon, Ahmed Ali, Younes Samih, Hamdy Mubarak, and James Glass. 2020.
\newblock Adi17: A fine-grained arabic dialect identification dataset.
\newblock In \emph{ICASSP 2020-2020 IEEE International Conference on Acoustics, Speech and Signal Processing (ICASSP)}, pages 8244--8248. IEEE.

\bibitem[{Sullivan et~al.(2023)Sullivan, Elmadany, and Abdul-Mageed}]{sullivan23_interspeech}
Peter Sullivan, AbdelRahim Elmadany, and Muhammad Abdul-Mageed. 2023.
\newblock \href {https://doi.org/10.21437/Interspeech.2023-1005} {On the robustness of arabic speech dialect identification}.
\newblock In \emph{Interspeech 2023}, pages 5326--5330.

\bibitem[{Talafha et~al.(2024)Talafha, Kadaoui, Magdy, Habiboullah, Chafei, El-Shangiti, Zayed, Alhamouri, Assi, Alraeesi et~al.}]{talafha2024casablanca}
Bashar Talafha, Karima Kadaoui, Samar~Mohamed Magdy, Mariem Habiboullah, Chafei~Mohamed Chafei, Ahmed~Oumar El-Shangiti, Hiba Zayed, Rahaf Alhamouri, Rwaa Assi, Aisha Alraeesi, and 1 others. 2024.
\newblock Casablanca: Data and models for multidialectal arabic speech recognition.
\newblock \emph{arXiv preprint arXiv:2410.04527}.

\bibitem[{Toyin et~al.(2023)Toyin, Djanibekov, Kulkarni, and Aldarmaki}]{toyin2023artst}
Hawau Toyin, Amirbek Djanibekov, Ajinkya Kulkarni, and Hanan Aldarmaki. 2023.
\newblock Artst: Arabic text and speech transformer.
\newblock In \emph{Proceedings of ArabicNLP 2023}, pages 41--51.

\bibitem[{Toyin et~al.({2025})Toyin, Marew, Alblooshi, Magdy, and Aldarmaki}]{toyin2025arvoicemultispeakerdatasetarabic}
Hawau Toyin, Rufael Marew, Humaid Alblooshi, Samar~M. Magdy, and Hanan Aldarmaki. {2025}.
\newblock \href {https://doi.org/{10.21437/Interspeech.2025-1550}} {{ArVoice: A Multi-Speaker Dataset for Arabic Speech Synthesis}}.
\newblock In \emph{{Interspeech 2025}}, pages {4808--4812}.

\bibitem[{Valk and Alum{\"a}e(2021)}]{valk2021voxlingua107}
J{\"o}rgen Valk and Tanel Alum{\"a}e. 2021.
\newblock Voxlingua107: a dataset for spoken language recognition.
\newblock In \emph{2021 IEEE Spoken Language Technology Workshop (SLT)}, pages 652--658. IEEE.

\bibitem[{Waheed et~al.(2023)Waheed, Talafha, Sullivan, Elmadany, and Abdul-Mageed}]{waheed2023voxarabica}
Abdul Waheed, Bashar Talafha, Peter Sullivan, Abdelrahim Elmadany, and Muhammad Abdul-Mageed. 2023.
\newblock Voxarabica: A robust dialect-aware arabic speech recognition system.
\newblock In \emph{Proceedings of ArabicNLP 2023}, pages 441--449.

\bibitem[{Zaghouani and Charfi(2018)}]{zaghouani2018}
Wajdi Zaghouani and Anis Charfi. 2018.
\newblock Arap-tweet: A large multi-dialect twitter corpus for gender, age and language variety identification.
\newblock In \emph{Proceedings of the Eleventh International Conference on Language Resources and Evaluation (LREC 2018)}, Miyazaki, Japan. European Language Resources Association (ELRA).

\bibitem[{Zaidan and Callison-Burch(2011)}]{zaidan2011}
Omar~F. Zaidan and Chris Callison-Burch. 2011.
\newblock The arabic online commentary dataset: An annotated dataset of informal arabic with high dialectal content.
\newblock In \emph{Proceedings of the 49th Annual Meeting of the Association for Computational Linguistics: Human Language Technologies}, pages 37--41, Portland, Oregon, USA. Association for Computational Linguistics.

\bibitem[{Zerrouki and Balla(2017)}]{tashkeela}
Taha Zerrouki and Amar Balla. 2017.
\newblock \href {https://doi.org/10.1016/j.dib.2017.01.011} {Tashkeela: Novel corpus of arabic vocalized texts, data for auto-diacritization systems}.
\newblock \emph{Data in Brief}, 11.

\end{thebibliography}


\end{document}